\documentclass[11pt]{article}

\usepackage{acl}

\usepackage{times}
\usepackage{latexsym}

\usepackage{xcolor}
\usepackage{booktabs}
\usepackage{cleveref}
\usepackage{siunitx}
\usepackage{needspace} 
\usepackage{caption}

\usepackage[show]{chato-notes}
\usepackage[T1]{fontenc}

\usepackage[utf8]{inputenc}

\usepackage{microtype}

\usepackage{inconsolata}

\usepackage{graphicx}
\usepackage{subcaption}

\graphicspath{{../}}

\title{P1SCO: Social Dimensions from a Perspectivist Lens}

\author{
  \textbf{Amanda Cercas Curry$^1$} \quad \textbf{Gianmarco De Francisci Morales$^2$} \quad \textbf{Luca Maria Aiello$^3$} \\
  $^1$Independent Researcher \quad $^2$CENTAI, Turin \quad $^3$IT University of Copenhagen \\
  \texttt{amanda.cercas@gmail.com} \quad \texttt{gdfm@acm.org} \quad \texttt{luai@itu.dk}
}

\begin{document}
\maketitle

\begin{abstract}
We introduce \textsc{P1SCO}, a dataset of social media comments collected from three distinct platforms, annotated according to ten social dimensions to capture the diversity of social interactions and perceptions.
The dataset is carefully disaggregated to allow analysis at the level of individual comment, annotator, and platform.
In addition to social dimension labels, we include rich metadata on the annotators, including demographics, Big Five personality profiles, and political affiliation.
This combination of comment-level annotations and annotator-level features enables nuanced analyses of how social perception varies across platforms, individual differences, and demographic factors.
By preserving the diversity of annotator perspectives, our dataset supports studies of inter- and intra-annotator agreement, the influence of personality and political orientation on social interpretation, and the cross-platform dynamics of social discourse.
We release the dataset to facilitate research in computational social science, social psychology, and natural language processing, promoting reproducibility and enabling further exploration of social dimension perception across heterogeneous online communities.
\end{abstract}


\section{Introduction}

\begin{figure}
    \centering
    \includegraphics[width=\linewidth]{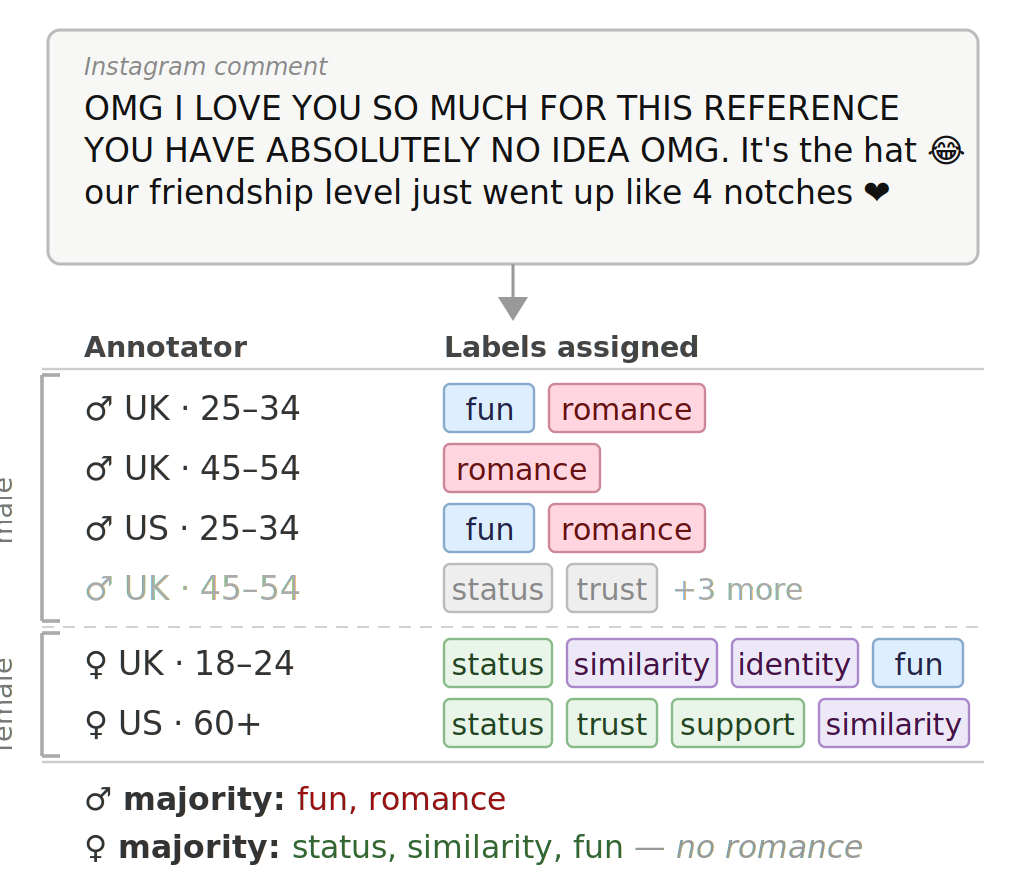}
    \caption{The same Instagram comment, interpreted through different demographic lenses. Male annotators assign romance and fun; female annotators assign status, similarity, and fun. Disagreement reflects genuine perspectival variation, not annotation error.}
    \label{fig:fig_1}
\end{figure}

In \textit{Ways of Seeing}, John Berger argues that we do not simply look at things---we interpret them through layers of culture, ideology, and personal experience.
What we ``see'' is already shaped by power, history, gender norms, class relations, and our own positionality.
The same thing can be said about language: if pragmatics is about meaning in context, then every individual brings their own context, shaping interpretation in ways as varied and layered as the perspectives Berger describes (see Fig \ref{fig:fig_1}).
Studying pragmatic expression and idiolects reveals not just how we speak, but how society itself is structured through interaction. 

The ambiguous nature of language and the idiosyncrasies of human label variation have inspired a growing body of literature studying the factors that affect different people's understanding of an utterance \cite[e.g.,][]{ovesdotter-alm-2011-subjective,hovy2021importance,plank-2022-problem}.
This work takes a perspectivist approach and acknowledges that linguistic expression is a performance of identity.

Previous work in computational social science has shown that conversations can be meaningfully analyzed along ten social dimensions that make up the fundamental building blocks of human relations~\citep{choi2020ten}.
These dimensions capture the social intent underlying utterances in a way that both interlocutors can \emph{perceive}.
Crucially, while these dimensions provide a structured framework for understanding social intent, their interpretation is far from uniform.
Each speaker brings a unique set of experiences, personality traits, and cultural background to an interaction, thus shaping how they express---and how they perceive---the same social intent.
This interplay between social dimensions and individual variation highlights the perspectivist nature of language: just as Berger observed that we see the world through layered lenses, we also enact and interpret social meaning through personal and situational filters.
By examining conversations through both these lenses, we can uncover patterns in how social goals are communicated and understood across diverse individuals.

By treating both social intent and individual variation as central to understanding language, we can begin to model how context and perspective jointly shape meaning in conversation and human relationships.
To this end, we introduce \textsc{P1SCO} (pronounced \textit{pisco}), Perspectives on the 10 Social dimensions of Conversation, a large dataset of social media comments from diverse platforms, manually labelled by diverse annotators.\footnote{\url{https://anonymous.4open.science/r/p1sco}}

\section{Background and Related Work}


Through an extensive review of the social science literature, prior work identified ten dimensions of the social pragmatics of language~\cite{deri2018coloring}.
These dimensions capture archetypal signals of social intent typically expressed in everyday conversations, such as conveying appreciation or offering emotional support.
\Cref{tab:definitions} reports the full set of dimensions along with their concise definitions.
Beyond their linguistic realization, these dimensions also correspond to core mental categories that individuals typically refer to when conceptualizing their social relationships.
As such, they provide a general theoretical framework for the study of social interaction across diverse contexts.
Accordingly, the ten dimensions framework has been widely adopted to investigate several types of social dynamics including agreement~\cite{monti2022language}, coordination~\cite{lucchini2022reddit, cava2023drivers}, and well-being~\cite{aiello2021epidemic, balsamo2023pursuit} in online communities, as well as to simulate human interaction dynamics with Large Language Models~\cite{breum2024persuasive}.

To operationalize these dimensions, prior work introduced a human-annotated corpus of social media text designed to train supervised natural language processing models for detecting signals of social pragmatics from conversations~\cite{choi2020ten}.
The corpus consists of \num{7855} sentences sampled from Reddit comments selected at random.
Each sentence contains between \num{6} and \num{20} words and includes at least one first- or second-person pronoun, to focus on phrases that follow a direct conversational structure.
Each sentence was annotated by three to five Amazon Mechanical Turk workers, who were allowed to assign multiple labels when applicable.
As the first dataset of its kind, this dataset is limited to a single online platform and is released in an aggregated form, reporting only total label counts per sentence and providing no information on annotator demographics.

\begin{table*}[t]
\centering

\resizebox{\textwidth}{!}{
\begin{tabular}{ll}
\toprule
\textbf{Dimension} & \textbf{Keywords} \\
\midrule
Knowledge & Exchange of ideas or information; learning, teaching: teaching, intelligence, competent, expertise, know-how, insight \\
Power & Having power over the behavior and outcomes of another: command, control, dominance, authority, pretentious, decisions \\
Status & Conferring status, appreciation, gratitude, or admiration upon another: admiration, appreciation, praise, thankful, respect, honor \\
Trust & Will of relying on the actions or judgments of another: trustworthy, honest, reliable, dependability, loyalty, faith \\
Support & Giving emotional or practical aid and companionship: friendly, caring, cordial, sympathy, companionship, encouragement \\
Romance & Intimacy among people with a sentimental or sexual relationship: love, sexual, intimacy, partnership, affection, emotional, couple \\
Similarity & Shared interests, motivations or outlooks: alike, compatible, equal, congenial, affinity, agreement \\
Identity & Shared sense of belonging to the same community or group: community, united, identity, cohesive, integrated \\
Fun & Experiencing leisure, laughter, and joy: funny, humour, playful, comedy, cheer, enjoy, entertaining \\
Conflict & Contrast or diverging views: hatred, mistrust, tense, disappointing, betrayal, hostile \\
\bottomrule
\end{tabular}
}
\caption{Definitions of the 10 social dimensions by \citet{choi2020ten}.}
\label{tab:definitions}
\end{table*}

Recent work in NLP has begun to argue that social meaning is inherently perspectival and that annotator disagreement may reflect legitimate interpretive variation rather than noise \cite{plank-2022-problem,cabitza2023toward}. 
Work in this area has thus focused on both sociodemographic \cite{pei2023annotator,hovy2021importance} and individual characteristics \cite{orlikowski2025beyond} to enrich models of linguistic phenomena.
A number of datasets have been published in recent years reflecting a perspectivist paradigm but these datasets are predominantly focused on offensiveness and sarcasm \cite{frenda2025perspectivist}.

Social media platforms differ substantially in demographics and interactional norms, leading to systematic variation in how content is produced, encountered, and interpreted.
According to the \citet{pew2025socialmedia}, Instagram has approximately 2 billion users worldwide and is especially popular among young adults in the United States, with 76\% of individuals aged 18--24 and 57\% of those aged 25--30 reporting use; platform usage is somewhat higher among women (55\%) than men (44\%).
YouTube, with an estimated 2.5 billion users globally, reaches one of the most demographically diverse audiences, exhibiting relatively balanced representation across gender, age groups, and racial categories.
Reddit, by contrast, serves a more demographically concentrated user base in the U.S., skewing younger than 40, predominantly male (29\% of men vs. 23\% of women in the U.S. use it), and disproportionately White or Asian.

These pronounced differences across platforms suggest that social meaning might be inherently perspectival: the same content may be perceived differently depending on platform context and the background of the individual interpreting it. 
Motivated by this insight, \textsc{P1SCO} adopts a perspectivist annotation framework that explicitly preserves annotator-level variation rather than collapsing judgments into a single ``ground truth.''
By combining cross-platform data with rich annotator metadata, the dataset enables systematic study of how social dimensions vary across platforms, how individual differences shape social perception, and how disagreement itself serves as a meaningful signal of social interpretation.

\section{Data Collection Methodology}

As a first step in the collection of our dataset, we source comments from social media.
To sample diverse comments, we follow previous work by \citet{choi2020ten}.
We randomly sample comments from three platforms: YouTube, Reddit, and Instagram.

\textbf{Instagram:}
We source posts, comments and replies from the Instagram Influencer Dataset \cite{kim2020multimodal}.
The dataset contains posts from over 38k influencers and over 1.6M comments.
We extract \num{20000} comments.

\textbf{YouTube:}
To randomly sample from YouTube, we follow the methodology proposed by \citet{zhou2011counting}.
We use the YouTube Search API and randomly generate prefixes to search for valid IDs.
We then use these IDs to randomly sample up to \num{100} comments from each video. We collect \num{20000} comments.

\textbf{Reddit:}
We randomly sample comments from the Pushshift Reddit dataset~\citep{baumgartner2020pushshift}.
We extract 0.01\% of the comments available between 2013 until 2024.

For every platform, we select first-level comments that are longer than \num{20} words and shorter than \num{100}, excluding emoji.
We exclude non-English comments.
We sample \num{2000} comments from those extracted for each platform, for a total of \num{6000} unique comments across all platforms.
While previous work selected only comments that included first- or second-person pronouns~\citep{choi2020ten}, to focus on direct interaction, we manually examined a sample of comments and found the filter to be unnecessary to obtain interactive comments.

\subsection{Annotation Interface}
We design an annotation interface using Streamlit.\footnote{\url{https://streamlit.io}}
Participants are first asked to answer a series of sociodemographic questions as well as complete the 10-question Big 5 personality questionnaire~\citep{RAMMSTEDT2007203}.
The participants are then guided through the definition of each dimension and some examples.
To source examples, we use the original dataset and find sentences where all annotators agreed on a single dimension.
Once the participants have read the instructions, they are shown the task.
Participants may select any of the ten dimensions as well as `Other', in which case they are prompted to specify.
They may also select `None'.
We use form validation to ensure `None' is not selected if other dimensions are.
To test the interface, we ran several pilots, first within the authors, then with small samples of participants.
Screenshots of the annotation interface are shown in the Appendix.

\section{Dataset}
We recruit participants from the U.K. and the U.S. using Prolific.\footnote{\url{https://www.prolific.com}}
To ensure data quality while allowing diverse views, we ensure that the demographic information provided through our study matches their Prolific profile.
We also remove users that look too long or too short to complete the task (5th and 95th percentile).
In addition, we collect three duplicate examples, using intra-annotator agreement as a further check \cite{abercrombie-etal-2025-consistency}.
Each annotator labeled \num{50} unique examples.
In total, after removing malicious participants, we collect \num{28775} annotations by \num{543} participants. 
Our participant pool is roughly even in terms of gender and nationality.
Sociodemographic distribution are shown in \Cref{fig:age_gender}.
A datasheet following \citet{gebru2021datasheets} can be found in the Appendix.

\begin{figure}
    \centering
    \includegraphics[width=1\linewidth]{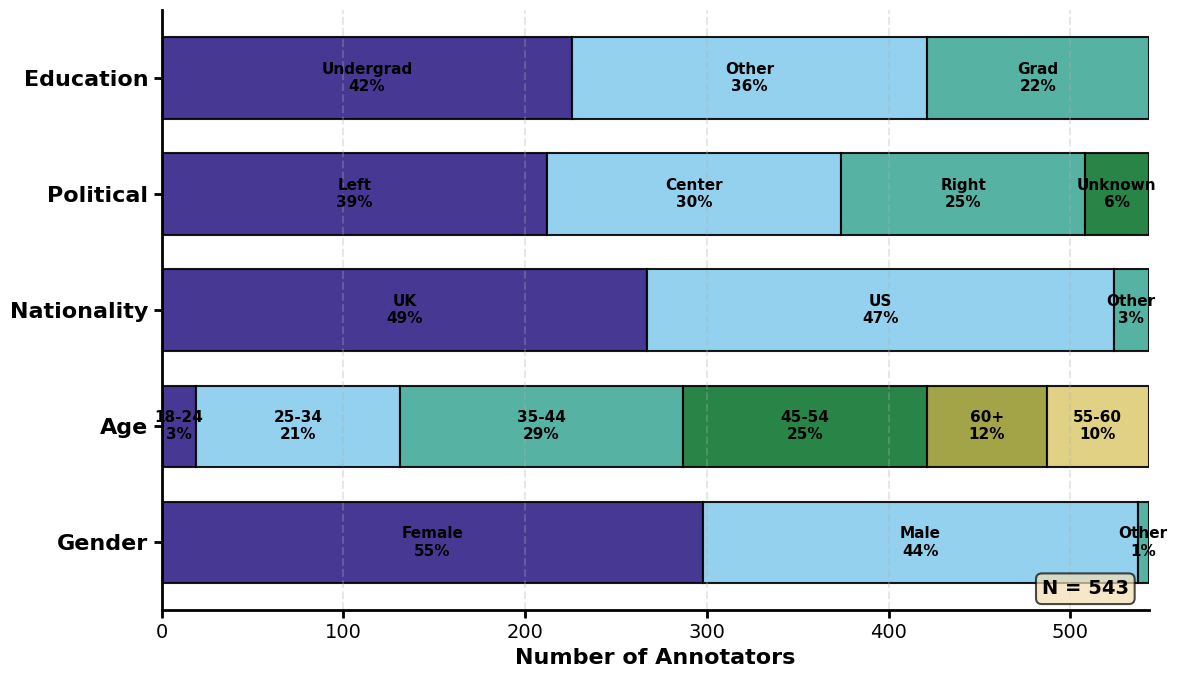}
    \caption{Demographic description of the participants.}
    \label{fig:age_gender}
\end{figure}

\subsection{Label Analysis}

\begin{figure}
    \centering
    \includegraphics[width=1\linewidth]{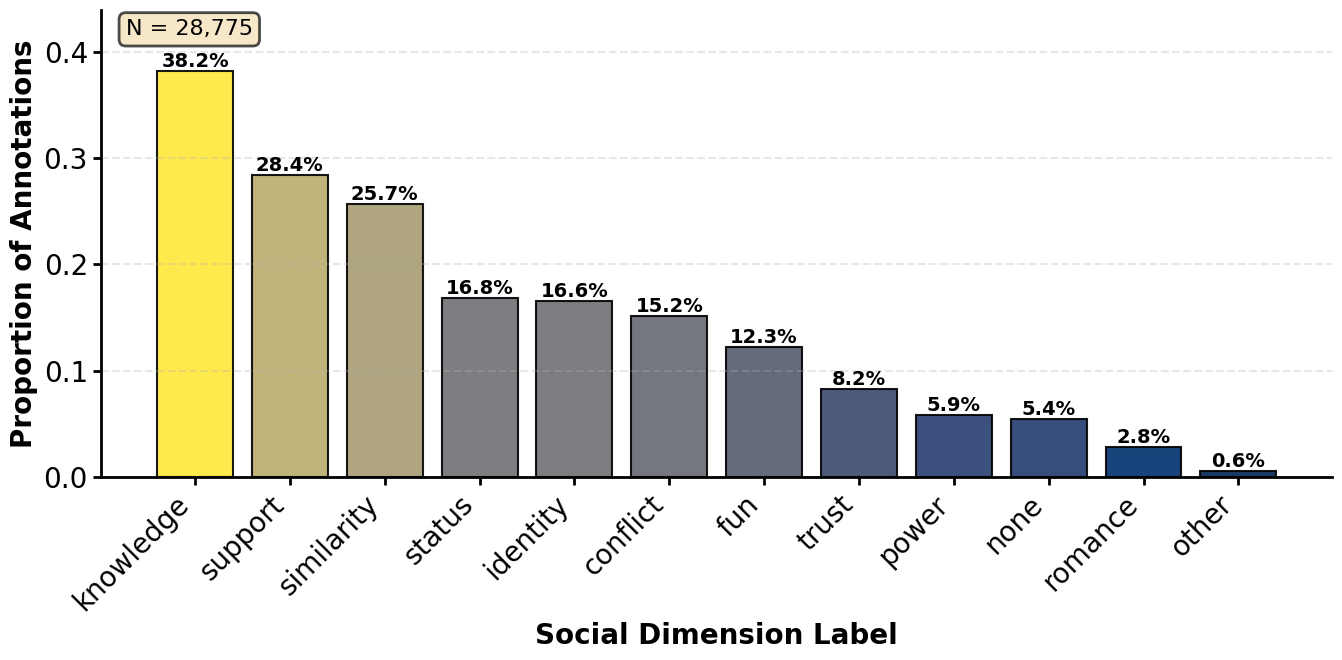}
    \caption{Overall distribution of assigned labels}
    \label{fig:overall_label_distribution}
\end{figure}

\Cref{fig:overall_label_distribution} shows the distribution of social dimension annotations across the dataset (N $= \num{28775}$).
Knowledge is the most frequently assigned label (38.2\%), followed by support (28.4\%), and similarity (25.7\%).
Mid-frequency dimensions include status, identity, and conflict; while trust, power, romance, and none are comparatively rare.
Overall, the distribution is highly skewed, with a small number of dimensions accounting for the majority of annotations.

\begin{figure}
    \centering
    \includegraphics[width=1\linewidth]{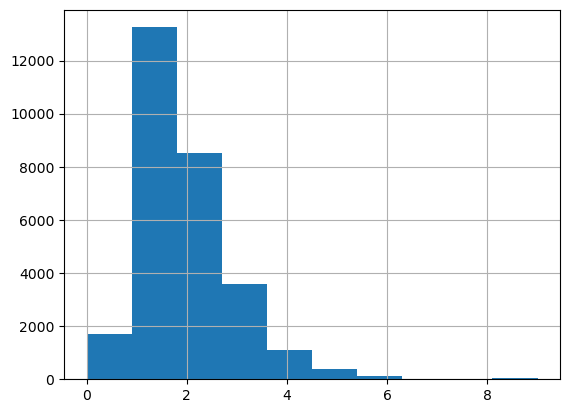}
    \caption{Distribution of number of labels per example.}
    \label{fig:dimensions_per_example}
\end{figure}

As our task allows for multiple labels, we consider how commonly dimensions co-occur.
\Cref{fig:dimensions_per_example} shows the distribution of the number of dimensions present in each example.
The majority of examples have at least two dimensions present: annotators frequently perceive multiple, overlapping social meanings within the same interaction.

\begin{figure*}[htbp]
    \centering
    \begin{subfigure}[b]{0.49\linewidth}
        \centering
        \includegraphics[width=\linewidth]{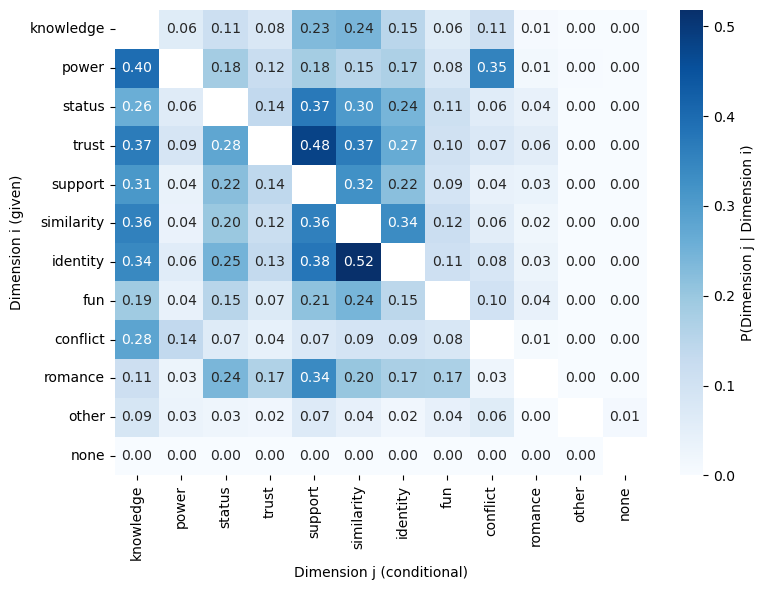}
        \caption{Conditional co-occurrence probabilities of social dimensions.}
        \label{fig:cond_probs}
    \end{subfigure}
    \hfill
    \begin{subfigure}[b]{0.48\linewidth}
        \centering
        \includegraphics[width=\linewidth]{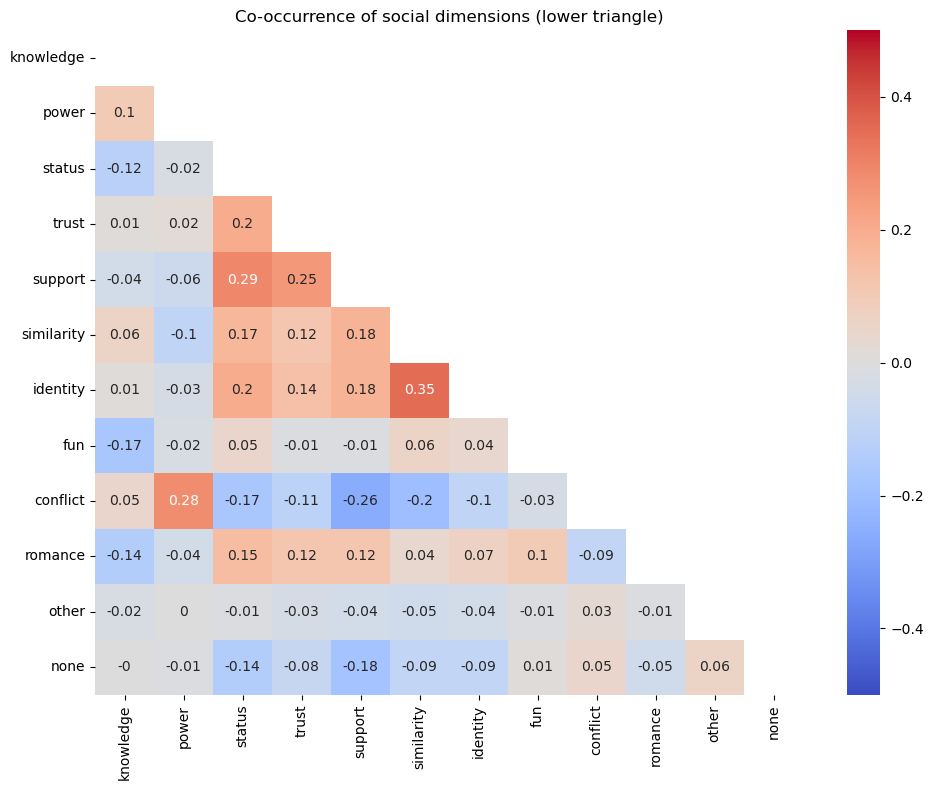}
        \caption{Joint co-occurrence probabilities of social dimensions.}
        \label{fig:cooccurrence}
    \end{subfigure}
    \caption{Co-occurrence matrices of social dimensions in the full dataset.}
    \label{fig:comparison_matrices}
\end{figure*}

\Cref{fig:comparison_matrices} shows the co-occurrence of labels and conditional probabilities across the entire dataset.
Three pairs of social dimensions tend to co-occur often: identity and similarity, trust and support, and power and conflict.
The first pair represents joint expressions of identity by the speakers, e.g., \emph{I have been the lone reader until recently when I found more reader people.}
The second pair indicates situations of mutual aid, sympathy, and empathy, e.g., \emph{@user awe. So kind. Thank you. I'm hoping that sharing my mom's journey helps others.}
The last pair is a typical example of a struggle, where an attempt to exert power elicits a conflictual reaction, e.g., \emph{Y'all need to stop posting straight up copying as an off brand. It's not a substitution for the original and it's not really funny or crappy, it's just an asshole thing to do}.
The figure reveals substantial asymmetry and structure in how dimensions co-occur.
Several dimensions (most notably knowledge, support, similarity, and identity) frequently co-occur with a wide range of other dimensions, whereas dimensions such as romance, power, and conflict tend to co-occur more selectively.
This pattern reinforces the view that social dimensions are not independent labels but interrelated interpretive constructs.

\begin{figure}
    \centering
    \includegraphics[width=1\linewidth]{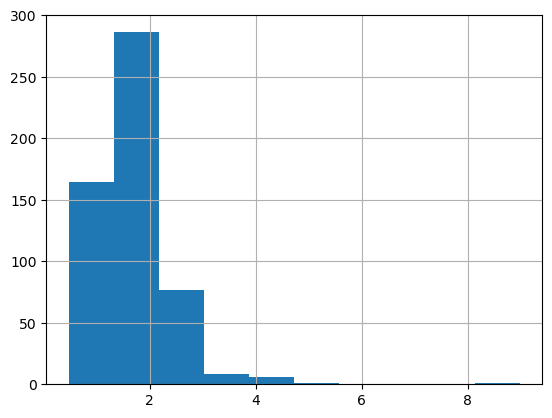}
    \caption{Distribution of average number of labels per annotator.}
    \label{fig:placeholder}
\end{figure}

\paragraph{Other} We manually look at the examples labeled `other'.
Generally, when asked to specify, users described either the emotion they perceived, sarcasm, or they were unable to understand the dimension from the context provided. 

\paragraph{Agreement} We measure agreement using pairwise Cohen's $\kappa$, shown in \Cref{tab:alpha_by_label,tab:alpha_gender_nationality}.
To understand the effects of sociodemographics, we calculate agreement stratifying by groups as well as overall.
We see moderate improvements in agreement, suggesting some group affinities, but agreement either remains stable or decreases in others, suggesting the task is highly subjective and likely challenging.
Pairwise Cohen's $\kappa$ is used over Krippendorff's $\alpha$ to capture localized agreement between annotators, which is more appropriate for a perspectival annotation task where systematic differences between annotators are expected and disagreement is not treated as noise.
We also report Krippendorff's \(\alpha\) in the Appendix.

\begin{table}[t]
\small
\centering

\begin{tabular}{lcccc}
\toprule
Label & Instagram & Reddit & YouTube & Average \\
\midrule
other & 0.992 & 0.976 & 0.971 & 0.980 \\
romance & 0.836 & 0.936 & 0.940 & 0.904 \\
none & 0.927 & 0.743 & 0.755 & 0.808 \\
power & 0.884 & 0.751 & 0.747 & 0.794 \\
trust & 0.585 & 0.774 & 0.765 & 0.708 \\
fun & 0.631 & 0.711 & 0.732 & 0.691 \\
conflict & 0.888 & 0.576 & 0.569 & 0.677 \\
status & 0.329 & 0.677 & 0.650 & 0.552 \\
identity & 0.383 & 0.518 & 0.568 & 0.490 \\
support & 0.190 & 0.545 & 0.511 & 0.415 \\
similarity & 0.214 & 0.391 & 0.454 & 0.353 \\
knowledge & 0.414 & 0.288 & 0.322 & 0.341 \\
\midrule
Average & 0.606 & 0.657 & 0.665 & 0.643 \\
\bottomrule
\end{tabular}
\caption{Pairwise Kappa by social dimension label and platform. Values represent the average pairwise kappa across all annotator pairs.}
\label{tab:alpha_by_label}
\end{table}

\begin{table*}[ht]

\centering
\resizebox{\textwidth}{!}{
\begin{tabular}{lccccccccccccccc}
\toprule
Label & \multicolumn{5}{c}{YouTube} & \multicolumn{5}{c}{Reddit} & \multicolumn{5}{c}{Instagram} \\
\cmidrule(lr){2-6}\cmidrule(lr){7-11}\cmidrule(lr){12-16}
 & Overall & Female & Male & US & UK & Overall & Female & Male & US & UK & Overall & Female & Male & US & UK\\
\midrule
knowledge & 0.322 & 0.323 & 0.314 & 0.309 & 0.318 & 0.288 & 0.316 & 0.245 & 0.289 & 0.280 & 0.414 & 0.434 & 0.413 & 0.390 & 0.411 \\
power & 0.747 & 0.758 & 0.749 & 0.739 & 0.780 & 0.751 & 0.745 & 0.755 & 0.752 & 0.756 & 0.884 & 0.883 & 0.876 & 0.868 & 0.896 \\
status & 0.650 & 0.674 & 0.582 & 0.619 & 0.674 & 0.677 & 0.684 & 0.645 & 0.615 & \textbf{0.741} & 0.329 & 0.361 & 0.298 & 0.311 & 0.368 \\
trust & 0.765 & 0.779 & 0.754 & 0.748 & 0.773 & 0.774 & 0.793 & 0.740 & 0.765 & 0.765 & 0.585 & 0.614 & 0.562 & 0.569 & 0.609 \\
support & 0.511 & 0.524 & 0.474 & 0.468 & 0.548 & 0.545 & 0.586 & 0.502 & 0.505 & 0.566 & 0.190 & 0.206 & 0.170 & 0.175 & 0.203 \\
similarity & 0.454 & 0.491 & 0.388 & 0.471 & 0.446 & 0.391 & 0.417 & 0.340 & 0.416 & 0.379 & 0.214 & 0.234 & 0.200 & 0.237 & 0.188 \\
identity & 0.568 & 0.603 & 0.510 & 0.555 & 0.584 & 0.518 & 0.552 & 0.466 & 0.545 & 0.500 & 0.383 & 0.403 & 0.361 & \textbf{0.437} & 0.356 \\
fun & 0.732 & 0.734 & 0.718 & 0.711 & 0.757 & 0.711 & 0.710 & 0.719 & 0.706 & 0.730 & 0.631 & 0.639 & 0.642 & 0.595 & 0.654 \\
conflict & 0.569 & 0.569 & 0.559 & 0.537 & 0.611 & 0.576 & 0.561 & 0.605 & 0.546 & 0.606 & 0.888 & 0.900 & 0.880 & 0.880 & 0.885 \\
other & 0.971 & 0.978 & 0.966 & 0.981 & 0.962 & 0.976 & 0.984 & 0.960 & 0.975 & 0.979 & 0.992 & 0.993 & 0.990 & 0.992 & 0.993 \\
none & 0.755 & 0.743 & 0.773 & 0.781 & 0.724 & 0.743 & 0.718 & 0.784 & 0.758 & 0.728 & 0.927 & 0.915 & 0.940 & 0.923 & 0.928 \\
romance & 0.940 & 0.949 & 0.929 & 0.926 & 0.963 & 0.936 & 0.937 & 0.934 & 0.914 & 0.954 & 0.836 & 0.854 & 0.802 & 0.830 & 0.839 \\
\bottomrule
\end{tabular}
}
\caption{Within-group agreement with pairwise Cohen's $\kappa$: gender and nationality by platform. Bold values indicate $\kappa$ substantially higher ($>0.05$) than overall agreement for that platform.}
\label{tab:alpha_gender_nationality}
\end{table*}

\paragraph{Mixed Effect Model}

We used mixed-effects logistic regression models to examine how annotator characteristics predict label assignment.
We fit separate models for each of the 10 social dimension labels (\emph{knowledge, power, status, trust,
support, similarity, identity, fun, conflict, romance; \emph{plus} other \emph{and} none}).

Fixed effects included demographic variables---gender (female vs.\ male/other), nationality (U.S. vs.\ U.K.), age group (reference category: most common group), education level (reference category: most common), and ethnicity (reference category: most common).
We collapse both the U.K. and U.S. political spectra into left, right, or center (reference category: most common). 
Platform is included as a categorical predictor with three levels (Instagram, YouTube, Reddit; reference category: Instagram).
In addition, we include the Big Five personality traits---Openness, Conscientiousness, Extraversion, Agreeableness, and Neuroticism---as continuous predictors and standardized (\(M = 0\), \(SD = 1\)) prior to analysis.

\begin{figure*}
    \centering
    \includegraphics[width=1\linewidth]{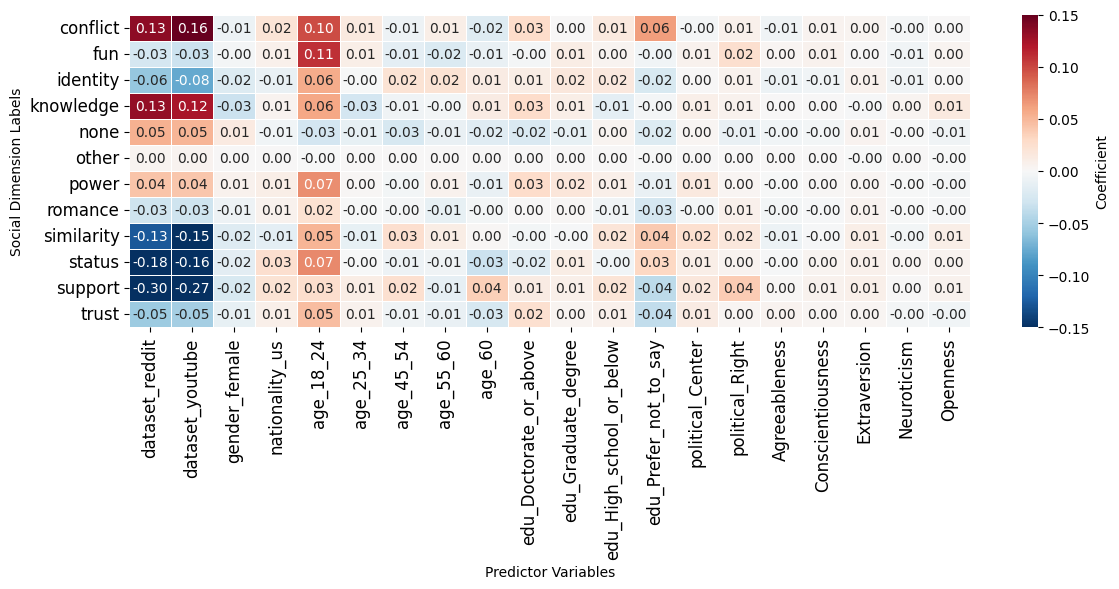}
    \caption{Coefficients for the mixed effects model. For the source platform, we use Instagram as a reference. For the other groups, we use the subgroup with the largest sample.}
    \label{fig:mixed_effects}
\end{figure*}

Models were fitted using the \texttt{mixedlm} function from the \texttt{statsmodels} package in Python, with maximum likelihood estimation (\texttt{REML = False}).
Statistical significance was assessed at \(\alpha = 0.05\).
The coefficients are shown in \Cref{fig:mixed_effects}.
The platform has the strongest effect, pointing to context and platform affordances and characteristics having the biggest effect on interactions.
Sociodemographics show modest effects, especially gender, age, and education. 

\begin{table}[hbp]
    \sisetup{table-number-alignment=right}
    \centering
    
    \begin{tabular}{lS[table-format=1.3]S[table-format=1.1]S[table-format=1.3]S[table-format=5.0]}
        \toprule
        Gender & \text{Mean} & \text{Median} & \text{Std. Dev.} & \text{Count} \\
        \midrule
        Female & 1.698 & 1.0 & 0.949 & 15788 \\
        Male   & 1.828 & 2.0 & 1.101 & 12669 \\
        \bottomrule
    \end{tabular}
    \caption{Descriptive statistics for the number of labels assigned by the annotators disaggregated by gender.}    
    \label{tab:gender_stats}
\end{table}

\begin{figure}[h]
    \centering
    \includegraphics[width=\linewidth]{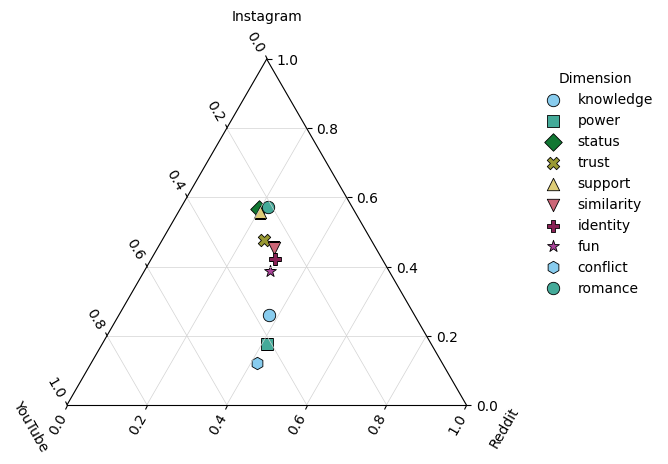}
    \caption{Proportion of each dimension per platform.}
    \label{fig:label_distribution}
\end{figure}

\begin{figure}
    \centering
    \includegraphics[width=\linewidth]{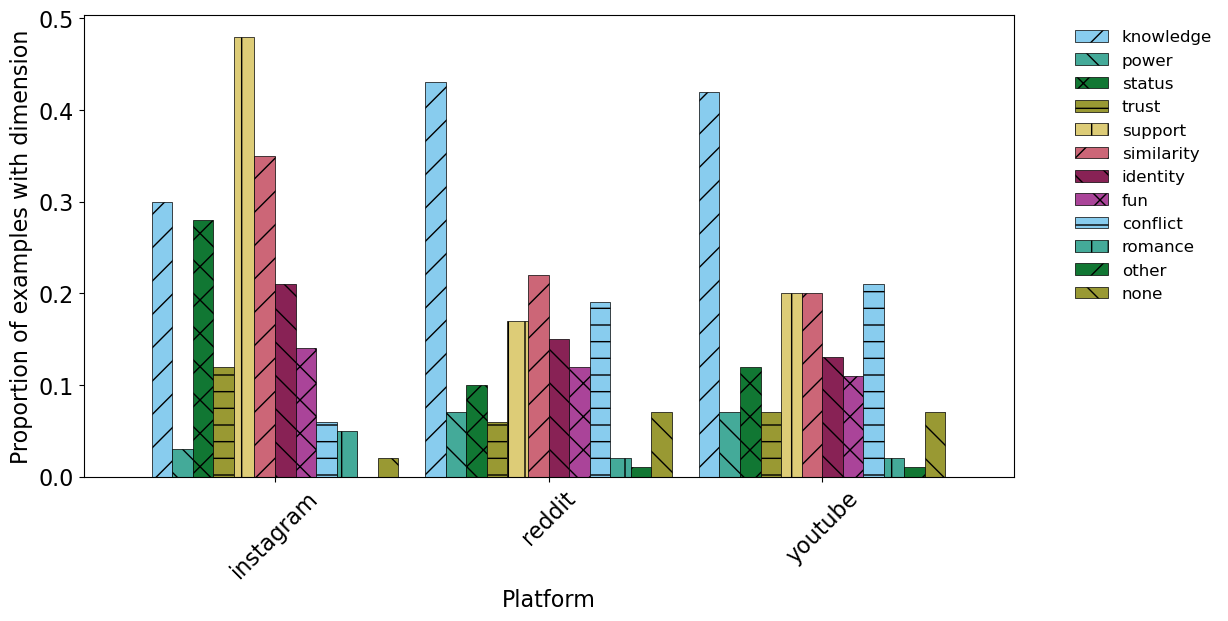}
    \caption{Distribution of presence of each dimension for each platform.}
    \label{fig:label_distribution_platform}
\end{figure}

\subsection{Platform Differences}
Given the differences in affordances and audience, we expect to see variation in the distribution of labels between platforms.
\Cref{fig:label_distribution} shows the label distributions for each platform. 
YouTube and Reddit show the most similar distributions: knowledge is the most dominant dimension present, followed by similarity, support, and conflict.
In contrast, Instagram interactions are dominated by support, similarity, knowledge, and status.
Romance remains low across platforms, but is significantly more common on Instagram.
Instagram also shows a significantly smaller proportion of posts with no dimensions present.
In contrast to Instagram, where support and similarity dominate, Reddit and YouTube are more strongly characterized by knowledge and conflict, pointing to different interactional expectations across platforms.
These differences reflect not only variation in platform interaction styles, but also how annotators interpret social meaning in context, with platform cues influencing which dimensions are perceived as salient.

\section{Sociodemographic Differences}
Given the findings of the previous model, we explore more in depth the sociodemographic differences in labels.
We find that annotators under 25, males, and US-nationals tend to assign a larger number of labels per example than other demographics groups.
In addition, gender and personality traits show effects on label distribution:

\subsection{Gender}
Debora Tannen presented \textit{genderlects}, the idea that men and women use language differently, with women using language to build rapport and men to show and gain power~\cite{tannen1990gender}.
While Tannen's original claims concern language production, we extend this logic to perception: if gendered socialisation shapes how one uses language, it plausibly shapes how one interprets it.
We therefore ask whether male and female annotators systematically differ in which social dimensions they perceive, a perceptual analogue of the genderlect hypothesis.

We first check whether men and women assign the same number of labels overall.
On average, men assign more labels per example than women (see \Cref{tab:gender_stats}).
This difference is statistically significant (Mann-Whitney-U test, $p < .001$). 
 
\begin{figure*}[htbp]
    \centering
    \begin{subfigure}[b]{0.48\linewidth}
        \centering
        \includegraphics[width=\linewidth]{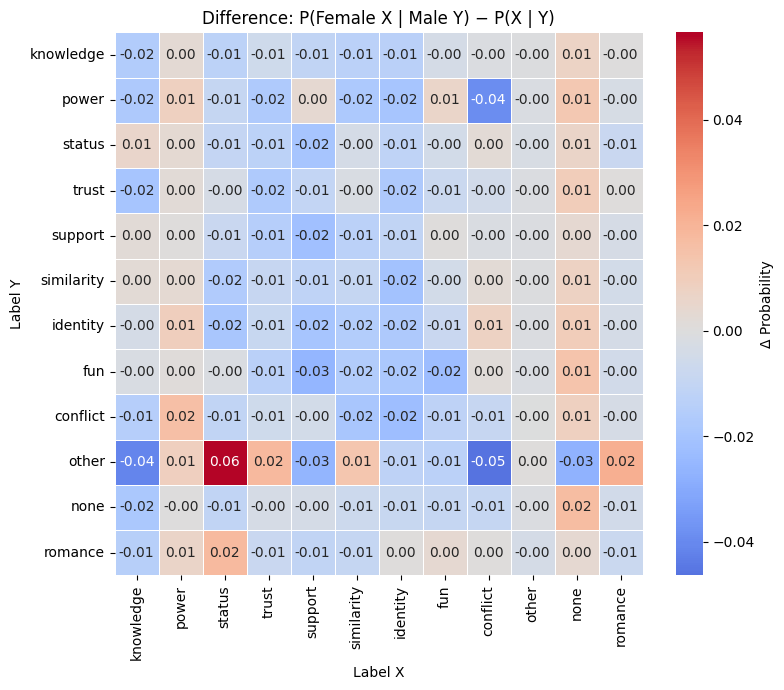}
        \caption{Conditional probabilities: how likely is a woman to label an example X given that a man labelled it Y}
        \label{fig:gender_probabilities}
    \end{subfigure}
    \hfill
    \begin{subfigure}[b]{0.48\linewidth}
        \centering
        \includegraphics[width=\linewidth]{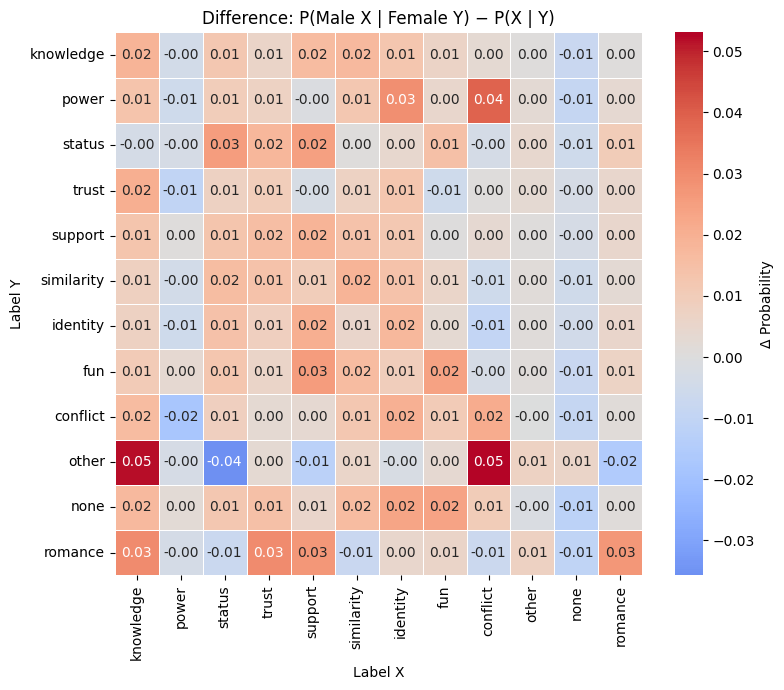}
        \caption{Conditional probabilities: how likely is a man to label an example X given that a woman labelled it Y}
        \label{fig:men_given_women_probabilities}
    \end{subfigure}
    \caption{Conditional probabilities between genders.}
    \label{fig:gender_conditional_probabilities}
\end{figure*}

A chi-square test comparing the distribution of applied labels by gender is significant, $\chi^2(10) = 143.33$, $p < .001$, indicating that men and women differed not only in the number of labels assigned but also in the relative frequency with which specific labels are applied.
This analysis conditions on the total number of labels assigned, isolating differences in label selection from overall labeling propensity.
When analyzing each platform, gender differences in label selection were larger on Reddit ($p < .001$) and YouTube ($p < .001$), while the effect on Instagram was weaker ($p = .033$).
These results suggest that gendered annotation strategies interact with platform-specific content characteristics rather than reflecting a uniform difference across platforms.

To understand how the labels differ, we calculate conditional co-occurrence probabilities (see \Cref{fig:gender_conditional_probabilities}).
The diagonal of the matrices shows overall agreement between men and women.
Notably, women often label an item as other when men assign status.
In turn, women are less likely to label something as power when men have labeled it as conflict or other.
Conversely, men are more likely to label other women's conflict as other or power, and to label romance where women see knowledge, trust, or support. 
 This pattern is directionally consistent with Tannen's characterization of male-coded and female-coded communicative orientations, now observed at the level of social perception rather than production. 
 However, the effect sizes are modest and interact with platform, suggesting that gendered interpretive schemas are real but contextually modulated.



\subsection{Big 5 Personality Traits}

\begin{figure}[ht!]
    \centering
    \includegraphics[width=1\linewidth]{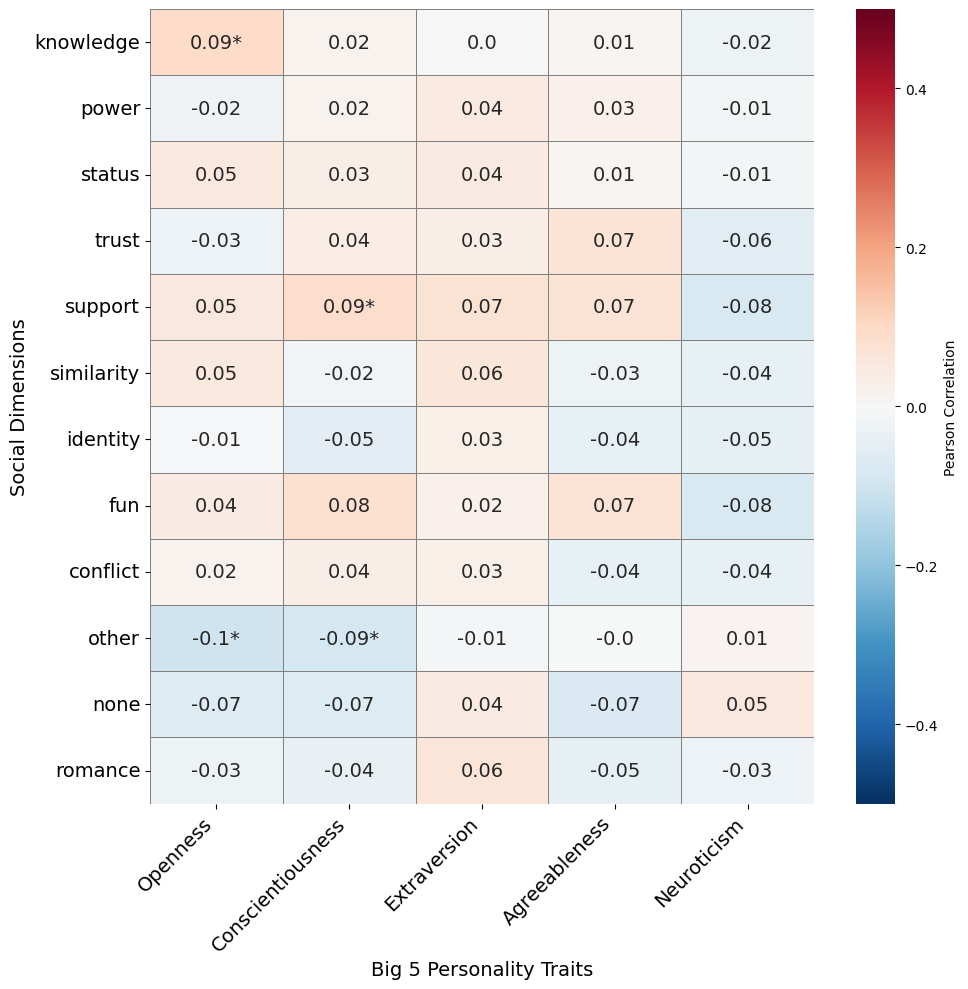}
    \caption{Correlations between Big 5 personality traits and label assignments for each dimension. Significance is Bonferroni-corrected.}
    \label{fig:personality_corr}
\end{figure}

\Cref{fig:personality_corr} shows the Pearson correlations between annotators' Big Five personality traits and the proportion of social dimensions they assigned.
Overall, correlations are small in magnitude, indicating no strong linear relationship between personality traits and annotation behavior.
A small number of weak but statistically significant associations are observed. For example, openness shows a modest positive correlation with knowledge annotations, while conscientiousness is weakly associated with support.
Conversely, the `other' category is negatively correlated with both openness and conscientiousness.
These patterns suggest that annotator characteristics may subtly influence which social dimensions are foregrounded, without implying systematic bias or determinism in annotation decisions.
Given the small effect sizes and correlational nature of the analysis, these results should be interpreted as exploratory rather than explanatory.


\section{Discussion and Conclusion}
This work provides empirical evidence for perspectivism in social dimension annotation, demonstrating that demographic characteristics, personality traits, and platform context systematically shape how annotators perceive and label social interactions.
Through mixed-effects modeling of over 28k annotations, we show that individual differences are a meaningful signal reflecting genuine diversity in social perception.

Our key contributions are:
we collect the first dataset of perspectivist unaggregated annotations of social dimensions; 
through this dataset, we demonstrate that demographic factors (gender, age, political ideology, nationality) and Big Five personality traits significantly predict label assignment across the 10 social dimensions, with effects consistent across Instagram, YouTube, and Reddit.
We show that homogeneous demographic groups exhibit higher within-group agreement than the overall population, supporting the premise that shared experiences create convergent interpretive frameworks;
we provide methodological guidance for perspectivist annotation, including preserving individual perspectives, modeling annotator characteristics explicitly, and reporting demographic composition transparently.

Our findings contribute to a growing body of work acknowledging human factors in computational humanities, and our dataset enables future work to develop new methods and models for human-centered studies. 
Future work should extend this framework to additional languages and cultures, explore intersectional demographic effects, and develop computational models that explicitly represent multiple viewpoints.
The dataset and code are available at \url{https://anonymous.4open.science/r/p1sco} to facilitate such research.

\section*{Limitations}
\paragraph{Annotation quality} Social dimensions are inherently interpretive, and annotations reflect annotators’ perceptions rather than objective properties of text or the dimensions intended by the commenters.
While this perspectival approach is a central contribution of the dataset, it also limits the extent to which labels can be treated as ground truth or used for definitive classification tasks.
Moreover, annotating ten possible dimensions is a cognitively difficult task. While annotators assigned social dimension labels, they did not provide explicit rationales for their choices.
This limits insight into the reasoning behind disagreements and constrains fine-grained analysis of interpretive variation.

\paragraph{Diversity} While we took care to expand our pool of annotators, our pool was not ethnically diverse and was focused on the United States and the United Kingdom.
As a result, the dataset reflects a bounded set of perspectives, and social meanings may be interpreted differently by annotators from other cultural or socio-linguistic contexts.
Future work could expand this pool to be more diverse and to cover languages other than English. 

\paragraph{Generalizability} The dataset consists of content from a small number of social media platforms, but our study shows platform has an important impact on the labels.
This may limit our findings generalizability to other social media platforms.

\paragraph{Label Distribution} Several social dimensions are relatively rare, leading to skewed label distributions.
This affects the reliability of agreement metrics and limits the usefulness of the dataset for training balanced supervised models without additional resampling or aggregation.

\section*{Ethical Considerations}
Annotators were paid the hourly wage recommended by the platform (9 GBP/hour).
While we took care to filter upsetting material (in particular, pornographic bots on Reddit), annotators were warned that the material they may see could be offensive in nature given the sources of data.
The dataset is available fully anonymized. 

 \section*{Acknowledgments}
AI-based assistants were used to provide coding, writing and editing assistance during manuscript preparation. All content, analyses, and conclusions were reviewed and verified by the authors.

\bibliography{aaai2026}

@inproceedings{baumgartner2020pushshift,
	author = {Baumgartner, Jason and Zannettou, Savvas and Keegan, Brian and Squire, Megan and Blackburn, Jeremy},
	booktitle = {Proceedings of the International AAAI Conference on Web and Social Media},
	pages = {830--839},
	title = {The pushshift reddit dataset},
	volume = {14},
	year = {2020}}

@inproceedings{pei2023annotator,
  title={When do annotator demographics matter? measuring the influence of annotator demographics with the POPQUORN dataset},
  author={Pei, Jiaxin and Jurgens, David},
  booktitle={Proceedings of the 17th linguistic annotation workshop (LAW-XVII)},
  pages={252--265},
  year={2023}
}

@inproceedings{choi2020ten,
	author = {Choi, Minje and Aiello, Luca Maria and Varga, Kriszti{\'a}n Zsolt and Quercia, Daniele},
	booktitle = {Proceedings of The Web Conference 2020},
	pages = {1514--1525},
	title = {Ten social dimensions of conversations and relationships},
	year = {2020}}

@inproceedings{kim2020multimodal,
	author = {Kim, Seungbae and Jiang, Jyun-Yu and Nakada, Masaki and Han, Jinyoung and Wang, Wei},
	booktitle = {Proceedings of The Web Conference 2020},
	pages = {2878--2884},
	title = {Multimodal Post Attentive Profiling for Influencer Marketing},
	year = {2020}}

@article{RAMMSTEDT2007203,
	author = {Beatrice Rammstedt and Oliver P. John},
	doi = {https://doi.org/10.1016/j.jrp.2006.02.001},
	issn = {0092-6566},
	journal = {Journal of Research in Personality},
	number = {1},
	pages = {203-212},
	title = {Measuring personality in one minute or less: A 10-item short version of the Big Five Inventory in English and German},
	url = {https://www.sciencedirect.com/science/article/pii/S0092656606000195},
	volume = {41},
	year = {2007}}

@misc{pew2025socialmedia,
	author = {{Pew Research Center}},
	howpublished = {\url{https://www.pewresearch.org/internet/fact-sheet/social-media/#who-uses-each-social-media-platform}},
	note = {Accessed: 2026-01-13},
	title = {Social Media Fact Sheet},
	year = {2025}}

@article{monti2022language,
  title={The language of opinion change on social media under the lens of communicative action},
  author={Monti, Corrado and Aiello, Luca Maria and De Francisci Morales, Gianmarco and Bonchi, Francesco},
  journal={Scientific Reports},
  volume={12},
  number={1},
  pages={17920},
  year={2022},
  publisher={Nature Publishing Group UK London}
}

@inproceedings{breum2024persuasive,
  title={The persuasive power of large language models},
  author={Breum, Simon Martin and Egdal, Daniel V{\ae}dele and Mortensen, Victor Gram and M{\o}ller, Anders Giovanni and Aiello, Luca Maria},
  booktitle={Proceedings of the International AAAI Conference on Web and Social Media},
  volume={18},
  pages={152--163},
  year={2024}
}

@article{cava2023drivers,
  title={Drivers of social influence in the Twitter migration to Mastodon},
  author={La Cava, Lucio and Aiello, Luca Maria and Tagarelli, Andrea},
  journal={Scientific Reports},
  volume={13},
  number={1},
  pages={21626},
  year={2023},
  publisher={Nature Publishing Group UK London}
}

@article{aiello2021epidemic,
  title={How epidemic psychology works on Twitter: Evolution of responses to the COVID-19 pandemic in the US},
  author={Aiello, Luca Maria and Quercia, Daniele and Zhou, Ke and Constantinides, Marios and {\v{S}}{\'c}epanovi{\'c}, Sanja and Joglekar, Sagar},
  journal={Humanities and social sciences communications},
  volume={8},
  number={1},
  year={2021},
  publisher={Springer Science and Business Media LLC}
}

@inproceedings{balsamo2023pursuit,
  title={The pursuit of peer support for opioid use recovery on reddit},
  author={Balsamo, Duilio and Bajardi, Paolo and Morales, Gianmarco De Francisci and Monti, Corrado and Schifanella, Rossano},
  booktitle={Proceedings of the international AAAI conference on web and social media},
  volume={17},
  pages={12--23},
  year={2023}
}

@article{lucchini2022reddit,
  title={From Reddit to Wall Street: The role of committed minorities in financial collective action},
  author={Lucchini, Lorenzo and Aiello, Luca Maria and Alessandretti, Laura and De Francisci Morales, Gianmarco and Starnini, Michele and Baronchelli, Andrea},
  journal={Royal Society Open Science},
  volume={9},
  number={4},
  pages={211488},
  year={2022},
  publisher={The Royal Society}
}

@article{deri2018coloring,
  title={Coloring in the links: Capturing social ties as they are perceived},
  author={Deri, Sebastian and Rappaz, Jeremie and Aiello, Luca Maria and Quercia, Daniele},
  journal={Proceedings of the ACM on Human-Computer Interaction},
  volume={2},
  number={CSCW},
  pages={1--18},
  year={2018},
  publisher={ACM New York, NY, USA}
}

@inproceedings{abercrombie-etal-2025-consistency,
    title = "Consistency is Key: Disentangling Label Variation in Natural Language Processing with Intra-Annotator Agreement",
    author = "Abercrombie, Gavin  and
      Dinkar, Tanvi  and
      Cercas Curry, Amanda  and
      Rieser, Verena  and
      Hovy, Dirk",
    editor = "Abercrombie, Gavin  and
      Basile, Valerio  and
      Frenda, Simona  and
      Tonelli, Sara  and
      Dudy, Shiran",
    booktitle = "Proceedings of the The 4th Workshop on Perspectivist Approaches to NLP",
    month = nov,
    year = "2025",
    address = "Suzhou, China",
    publisher = "Association for Computational Linguistics",
    url = "https://aclanthology.org/2025.nlperspectives-1.6/",
    doi = "10.18653/v1/2025.nlperspectives-1.6",
    pages = "63--74",
    ISBN = "979-8-89176-350-0"
}

@article{frenda2025perspectivist,
  title={Perspectivist approaches to natural language processing: a survey},
  author={Frenda, Simona and Abercrombie, Gavin and Basile, Valerio and Pedrani, Alessandro and Panizzon, Raffaella and Cignarella, Alessandra Teresa and Marco, Cristina and Bernardi, Davide},
  journal={Language Resources and Evaluation},
  volume={59},
  number={2},
  pages={1719--1746},
  year={2025},
  publisher={Springer}
}

@inproceedings{plank-2022-problem,
    title = "The ``Problem'' of Human Label Variation: On Ground Truth in Data, Modeling and Evaluation",
    author = "Plank, Barbara",
    editor = "Goldberg, Yoav  and
      Kozareva, Zornitsa  and
      Zhang, Yue",
    booktitle = "Proceedings of the 2022 Conference on Empirical Methods in Natural Language Processing",
    month = dec,
    year = "2022",
    address = "Abu Dhabi, United Arab Emirates",
    publisher = "Association for Computational Linguistics",
    url = "https://aclanthology.org/2022.emnlp-main.731/",
    doi = "10.18653/v1/2022.emnlp-main.731",
    pages = "10671--10682"
}

@inproceedings{cabitza2023toward,
  title={Toward a perspectivist turn in ground truthing for predictive computing},
  author={Cabitza, Federico and Campagner, Andrea and Basile, Valerio},
  booktitle={Proceedings of the AAAI Conference on Artificial Intelligence},
  volume={37},
  number={6},
  pages={6860--6868},
  year={2023}
}

@inproceedings{hovy2021importance,
  title={The importance of modeling social factors of language: Theory and practice},
  author={Hovy, Dirk and Yang, Diyi},
  booktitle={Proceedings of the 2021 Conference of the North American Chapter of the Association for Computational Linguistics: Human language technologies},
  pages={588--602},
  year={2021}
}

@article{orlikowski2025beyond,
  title={Beyond Demographics: Fine-tuning Large Language Models to Predict Individuals' Subjective Text Perceptions},
  author={Orlikowski, Matthias and Pei, Jiaxin and R{\"o}ttger, Paul and Cimiano, Philipp and Jurgens, David and Hovy, Dirk},
  journal={CoRR},
  year={2025}
}

@inproceedings{tannen1990gender,
  title={Gender differences in conversational coherence: Physical alignment and topical cohesion.},
  author={Tannen, Deborah},
  booktitle={This chapter is a slightly revised and shortened version of a paper presented at the" Gender Differences in Conversational Interaction" panel at the 1988 Georgetown University Round Table on Languages and Linguistics, Washington, DC, Mar 1988.},
  year={1990},
  organization={Ablex Publishing}
}

@inproceedings{zhou2011counting,
  title={Counting youtube videos via random prefix sampling},
  author={Zhou, Jia and Li, Yanhua and Adhikari, Vijay Kumar and Zhang, Zhi-Li},
  booktitle={Proceedings of the 2011 ACM SIGCOMM conference on Internet measurement conference},
  pages={371--380},
  year={2011}
}

@article{gebru2021datasheets,
  title={Datasheets for datasets},
  author={Gebru, Timnit and Morgenstern, Jamie and Vecchione, Briana and Vaughan, Jennifer Wortman and Wallach, Hanna and Iii, Hal Daum{\'e} and Crawford, Kate},
  journal={Communications of the ACM},
  volume={64},
  number={12},
  pages={86--92},
  year={2021},
  publisher={ACM New York, NY, USA}
}

@inproceedings{ovesdotter-alm-2011-subjective,
    title = "Subjective Natural Language Problems: Motivations, Applications, Characterizations, and Implications",
    author = "Ovesdotter Alm, Cecilia",
    editor = "Lin, Dekang  and
      Matsumoto, Yuji  and
      Mihalcea, Rada",
    booktitle = "Proceedings of the 49th Annual Meeting of the Association for Computational Linguistics: Human Language Technologies",
    month = jun,
    year = "2011",
    address = "Portland, Oregon, USA",
    publisher = "Association for Computational Linguistics",
    url = "https://aclanthology.org/P11-2019/",
    pages = "107--112"
}

\appendix

\section{Datasheet for the P1SCO Dataset}

\subsection*{Motivation}
Language is inherently ambiguous with different groups and individuals bring their own flavor to it.
This dataset aims to enable researchers to study diversity in the expression and perception of pragmatics.

\subsection*{Composition}

The dataset contains 28k annotated text-only comments from Instagram, YouTube and Reddit. 
The dataset is labeled with the ten social dimensions, other or none, as well as sociodemographic profiles and personality of the annotators.
We do not know the demographic profiles of the creators of the data that was annotated.
See main body of paper for more detailed demographics of the annotators.

\subsection*{Collection Process}
Described in the paper.
Crowdworkers are paid the wage recommended by Prolific of 9 GBP/hour.
We based the time estimate on in-house pilots completed by the researchers.
We adjusted the payment to crowdworkers to match the recommended payment.

\subsection*{Annotation Process}
Described in the paper.
Annotation interface shown in Appendix \ref{app:interface}.

\subsection*{Uses}
The intended use of this dataset is to study and model diversity of perspectives in pragmatics, both at the individual and group level.
The dataset will also enable the development of new methods for perspectivist models and frameworks.
The dataset may also be used to develop models that enable researchers to study social interactions over social media at a more fine-grained detail and with diverse perspectives and to benchmark models against diverse perspectives. 

\subsection*{Distribution}
The dataset is available at \url{https://anonymous.4open.science/r/p1sco/README.md} under an MIT license. The dataset should not be used to profile or otherwise violate the rights to dignity and autonomy of individuals or groups. 

\subsection*{Maintenance}
Anonymized for submission. 

\subsection*{Ethical Considerations and Limitations}
The main limitations are discussed in the paper. In terms of ethical considerations, the dataset has been anonymized by removing usernames and re-assigning worker IDs to crowdworkers. 
The dataset is mainly annotated by U.S. and U.K. citizens, who are mostly white.
In addition, the dataset labels show the perspective of a listener, not of the speaker.

\newpage
\onecolumn
\section{Agreement}
\label{app:agreement}

\begin{table*}[hbt!]
\caption{Within-group agreement: gender and nationality by platform. Bold values indicate $\alpha$ substantially higher ($>0.05$) than overall agreement for that dataset.}
\label{tab:alpha_gender_nationality}
\centering
\resizebox{\textwidth}{!}{
\begin{tabular}{lccccccccccccccc}
\toprule
Label & \multicolumn{5}{c}{YouTube}  & \multicolumn{5}{c}{Reddit}  & \multicolumn{5}{c}{Instagram} \\
\cmidrule(lr){2-6}\cmidrule(lr){7-11}\cmidrule(lr){12-16}
 & Overall & Female & Male & US & UK & Overall & Female & Male & US & UK & Overall & Female & Male & US & UK \\
\midrule
knowledge       & 0.30 & 0.31 & 0.29 & 0.31 & 0.29 & 0.27 & 0.28 & 0.24 & 0.26 & 0.27 & 0.30 & 0.31 & 0.31 & 0.27 & 0.31 \\
power           & 0.08 & 0.10 & 0.03 & 0.11 & 0.08 & 0.08 & \textbf{0.13} & 0.04 & 0.09 & 0.09 & 0.04 & 0.04 & 0.02 & 0.04 & 0.04 \\
status          & 0.17 & 0.17 & 0.13 & 0.15 & 0.18 & 0.12 & 0.14 & 0.08 & 0.06 & 0.16 & 0.16 & 0.19 & 0.13 & 0.15 & 0.18 \\
trust           & 0.05 & 0.06 & 0.07 & 0.05 & 0.03 & 0.06 & 0.05 & 0.07 & 0.02 & 0.09 & 0.02 & 0.03 & 0.00 & 0.00 & 0.04 \\
support         & 0.26 & 0.26 & 0.24 & 0.23 & 0.26 & 0.22 & 0.25 & 0.20 & 0.19 & 0.26 & 0.20 & 0.21 & 0.17 & 0.18 & 0.21 \\
similarity      & 0.15 & 0.17 & 0.12 & 0.16 & 0.14 & 0.12 & 0.13 & 0.11 & 0.10 & 0.13 & 0.14 & 0.15 & 0.15 & 0.15 & 0.12 \\
identity        & 0.09 & 0.08 & 0.12 & 0.08 & 0.13 & 0.09 & 0.08 & 0.08 & 0.07 & 0.09 & 0.07 & 0.09 & 0.03 & 0.06 & 0.07 \\
fun             & 0.30 & 0.29 & 0.29 & 0.31 & 0.30 & 0.29 & 0.28 & 0.29 & 0.27 & 0.32 & 0.26 & 0.27 & 0.24 & 0.23 & 0.27 \\
conflict        & 0.36 & 0.36 & 0.34 & 0.36 & 0.38 & 0.32 & 0.27 & 0.36 & 0.32 & 0.32 & 0.44 & 0.47 & 0.43 & 0.41 & 0.43 \\
romance         & 0.13 & 0.05 & 0.16 & 0.12 & 0.16 & 0.18 & 0.21 & 0.20 & 0.09 & \textbf{0.26} & 0.12 & 0.12 & 0.08 & 0.15 & 0.14 \\
other           & 0.02 & -0.01 & -0.01 & -0.00 & 0.04 & -0.01 & -0.00 & -0.01 & -0.00 & -0.01 & -0.00 & -0.00 & -0.00 & -0.00 & -0.00 \\
none            & 0.06 & 0.07 & 0.04 & 0.05 & 0.07 & 0.04 & 0.05 & 0.03 & 0.02 & 0.05 & 0.03 & 0.05 & 0.01 & 0.03 & \textbf{0.10} \\
\bottomrule
\end{tabular}
}
\end{table*}

\newpage
\onecolumn
\section{Data collection interface} \label{app:interface}

\begin{center}
    \includegraphics[width=\linewidth]{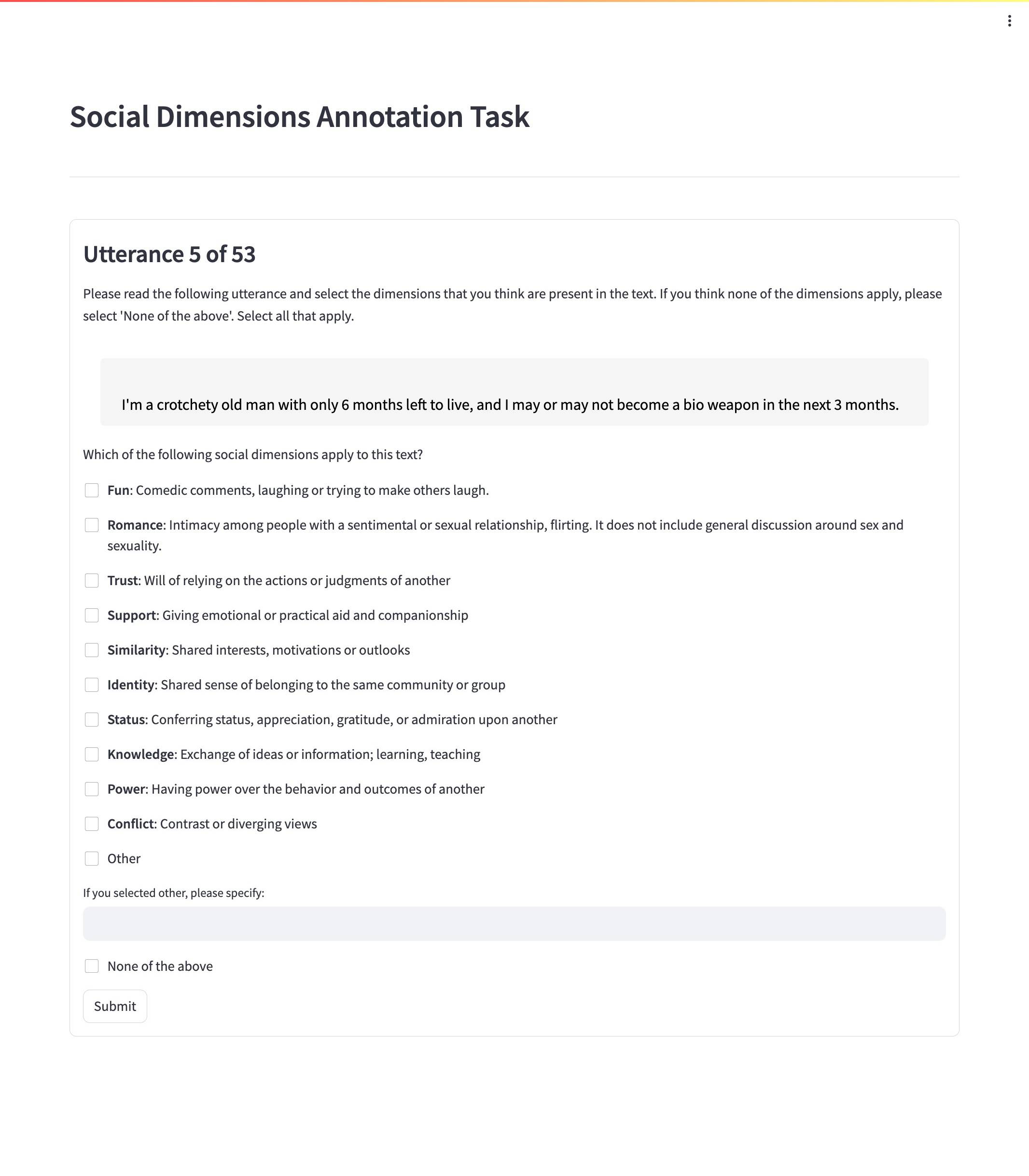}
\end{center}
\captionof{figure}{Screenshot of the P1SCO annotation interface. Participants were shown one utterance at a time and asked to select all applicable social dimensions. A free-text ``Other'' field and a ``None of the above'' option were also provided.}
\label{fig:placeholder}

\end{document}